# Hệ thống dẫn đường trong nhà cho người khiếm thị dựa trên công nghệ định vị UWB và giải thuật lập quỹ đạo D* Lite

# An Indoor Navigation System for the Visually Impaired based on UWB Positioning and D* Lite Path Planning Algorithm

Võ Chí Thành, Trần Lê Thăng Đồng, Lê Hoàng Minh Huy, Hà Mỹ Duyên,
Phạm Anh Tuấn, Đặng Thanh Hải, Trần Thuận Hoàng*

Viện Công nghệ Kỹ thuật Hàng không Vũ trụ, Đại học Duy Tân, Đà Nẵng
Email: tranthuanhoang@duytan.edu.vn

## Abstract
This paper proposes an indoor navigation system for the visually impaired, leveraging Ultra-Wideband (UWB) positioning technology and the D*Lite path planning algorithm. The system utilizes UWB sensors to provide precision localization in GPS-denied environments. The D* Lite algorithm is integrated to optimize travel trajectories and ensure rapid route re-planning in the presence of dynamic obstacles. Experimental results demonstrate that the system operates reliably with low latency, providing safety and flexibility for users in complex indoor spaces.

## Keywords
D* Lite, Ultra-Wideband, GPS-denied, localization, indoor navigation.

## Tóm tắt
Bài báo đề xuất một hệ thống dẫn đường trong nhà dành cho người khiếm thị dựa trên sự kết hợp giữa công nghệ định vị Ultra-Wideband (UWB) và giải thuật lập kế hoạch đường đi D* Lite. Hệ thống sử dụng các cảm biến UWB để xác định vị trí người dùng với độ chính xác cao trong môi trường không có tín hiệu GPS. Giải thuật D* Lite được tích hợp để tối ưu hóa quỹ đạo di chuyển và có khả năng tái lập lộ trình nhanh chóng khi xuất hiện các vật cản động trong không gian cố định. Kết quả thực nghiệm cho thấy hệ thống hoạt động ổn định, đảm bảo tính an toàn và linh hoạt cho người dùng trong các không gian trong nhà có nội thất được bố trí phức tạp.

## Symbols

| | |
|---|---|
| **D*** | D* Lite |

## Abbreviations

| | |
|---|---|
| LPA* | Lifelong Planning A* |
| *Rhs* | right-hand side |
| UWB | Ultra-Wideband |

## 1. Introduction

Independent and safe navigation in indoor environments remains one of the most significant challenges for visually impaired individuals. While outdoor positioning systems have advanced substantially based on GPS technology, these signals are considerably attenuated within indoor spaces and fail to meet the required precision [1][2][3]. Traditional assistive methods, such as white canes or guide dogs, only serve as passive reactions to nearby obstacles and cannot provide a comprehensive navigation solution from a starting point to a destination in complex environments like public buildings or hospitals. Therefore, the development of an intelligent positioning and navigation system capable of real-time interaction is of paramount importance.

To address the problem of accurate indoor localization, various wireless technologies have been deployed, including Wi-Fi, Bluetooth Low Energy (BLE), and ZigBee. However, these technologies are often based on Received Signal Strength Indication (RSSI), which is highly susceptible to interference and inaccuracies due to multi-path effects in confined spaces. In this study, Ultra-Wideband (UWB) technology is selected due to its wide bandwidth operation, allowing for high-precision Time of Flight (ToF) measurements. Compared to Wi-Fi or BLE, UWB provides centimeter-level positioning accuracy, low latency, and superior noise resistance, which are key factors in ensuring absolute safety for visually impaired users during movement [4][5][6][7]. Table I presents a comparative overview of common indoor positioning technologies. As shown, RSSI-based methods such as Wi-Fi, BLE, and ZigBee suffer from limited accuracy and poor multipath resistance, making them unsuitable for safety-critical applications. In contrast, UWB offers centimeter-level precision and high robustness against interference, which are essential requirements for guiding visually impaired users in complex indoor environments.

In addition to localization, the ability to plan travel trajectories (path planning) plays a decisive role in the system's effectiveness. In practice, an indoor environment is not an absolutely static space; the appearance of dynamic obstacles or changes in furniture positions requires the system to possess immediate reaction capabilities. The D* Lite algorithm is applied in this research due to its outstanding advantages in incremental re-planning. Unlike the traditional A* algorithm, which must recalculate the entire map when changes occur, D* Lite only updates the affected portions based on information from previous calculation steps. This not only optimizes computational resources on embedded devices but also ensures continuous navigation, enabling users to promptly handle arising situations in real-world environments [8][9][10][11].

TABLE I

| Technology | Accuracy | Latency | Bandwidth | Multi-path Resistance | Cost |
|---|---|---|---|---|---|
| UWB | 10 - 30 cm | Low | 500 MHz - 7.5 GHz | High | Medium |
| Wi-Fi (RSSI) | 1 - 5 m | Medium | 20 - 40 MHz | Low | Low |
| Bluetooth Low Energy | 1 - 3 m | Low | 2 MHz | Low | Very Low |
| ZigBee | 1 - 3 m | Low | 2 MHz | Low | Low |
| Ultrasound | < 10 cm | Very Low | | Very High | High |

Recent advancements in sensor fusion and path planning have significantly improved indoor navigation for the visually impaired [12][13][14]. This paper introduces an innovative head-mounted device integrating BNO055 IMU and DWM1000 modules to achieve precise localization and orientation. The D* Lite algorithm is utilized for efficient path planning and dynamic obstacle avoidance. By incorporating real-time voice prompts, the system provides a comprehensive, user-friendly solution for independent mobility. The paper is organized as follows: Section 2 details the system design describe the hardware architecture, the localization algorithms, grid mapping, and D* Lite implementation; Section 3 presents experimental results and feasibility analysis; finally, Section 4 concludes the study and suggests future directions.

## 2. System Design

To facilitate obstacle-free navigation for visually impaired users in unfamiliar environments, the system requires high-precision indoor localization and orientation data. The proposed solution employs a helmet equipped with a BNO055 IMU for accurate heading and a DWM1000 UWB module for trilateration-based positioning. A Raspberry Pi 4 serves as the central controller, executing pathfinding algorithms and delivering real-time audio guidance via a speaker, while a push-button allows for destination selection. To ensure ergonomic comfort and minimize helmet weight, the processing unit and battery are housed in a separate waist-worn module. Figure 2 illustrates the overall design of the system.

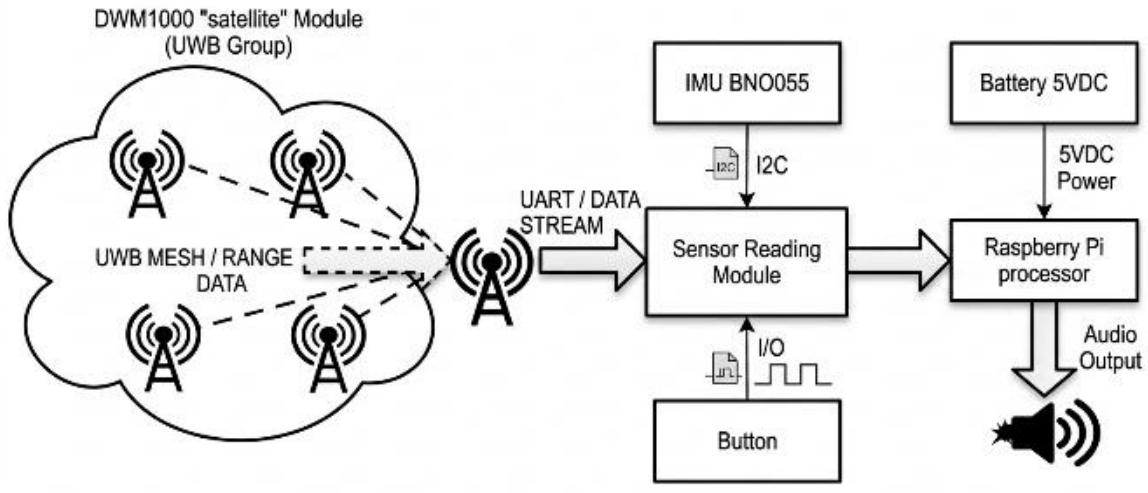

**Fig 1.** *System architecture*

### 2.1 Indoor positioning based on UWB

Indoor localization is achieved by deploying a network of DWM1000 anchors at fixed, predetermined coordinates. As the user moves, the head-mounted DWM1000 tag measures the distances to these anchors via Ultra-Wideband (UWB) ranging. This data is then transmitted to the central processor to compute the user's real-time coordinates using trilateration algorithms. While a minimum of three anchors is mathematically sufficient for 2D positioning, this system utilizes four anchors strategically placed at the corners of the environment (as illustrated in Figure 2) to enhance spatial resolution and measurement accuracy.

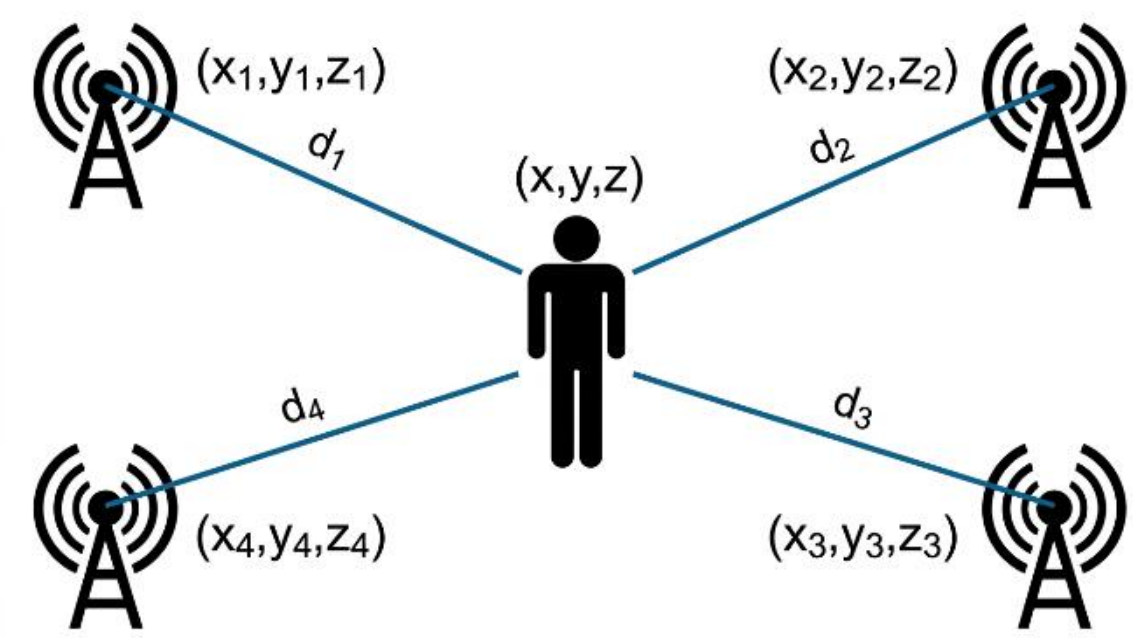

**Fig 2**. *Placement of satellites modules*

Each satellite has fixed coordinates $X_i = [x_i \quad y_i \quad z_i]$, the object to be located has coordinates $X = [x \quad y \quad z]$ and the distance from the object to the satellites is $d_i$ calculated as follows $(x-x_i)^2 + (y-y_i)^2 + (z-z_i)^2 = d_i^2$. Deploying for each sensor we have:

$$(x-x_1)^2 + (y-y_1)^2 + (z-z_1)^2 = d_1^2 \quad (1)$$
$$(x-x_2)^2 + (y-y_2)^2 + (z-z_2)^2 = d_2^2 \quad (2)$$
$$(x-x_3)^2 + (y-y_3)^2 + (z-z_3)^2 = d_3^2 \quad (3)$$
$$(x-x_4)^2 + (y-y_4)^2 + (z-z_4)^2 = d_4^2 \quad (4)$$

Take (2) minus (1), (3) minus (2), (4) minus (3) and (1) minus (4) respectively, we get:

$$\begin{aligned} 2[x(x_1-x_2)+y(y_1-y_2)+z(z_1-z_2)] &= d_2^2-(x_2^2+y_2^2+z_2^2)-d_1^2+(x_1^2+y_1^2+z_1^2)\\ 2[x(x_2-x_3)+y(y_2-y_3)+z(z_2-z_3)] &= d_3^2-(x_3^2+y_3^2+z_3^2)-d_2^2+(x_2^2+y_2^2+z_2^2)\\ 2[x(x_3-x_4)+y(y_3-y_4)+z(z_3-z_4)] &= d_4^2-(x_4^2+y_4^2+z_4^2)-d_3^2+(x_3^2+y_3^2+z_3^2)\\ 2[x(x_4-x_1)+y(y_4-y_1)+z(z_4-z_1)] &= d_1^2-(x_1^2+y_1^2+z_1^2)-d_4^2+(x_4^2+y_4^2+z_4^2) \end{aligned} \quad (5)$$

$$2\begin{bmatrix} x_1-x_2 & y_1-y_2 & z_1-z_2\\ x_2-x_3 & y_2-y_3 & z_2-z_3\\ x_3-x_4 & y_3-y_4 & z_3-z_4\\ x_4-x_1 & y_4-y_1 & z_4-z_1 \end{bmatrix}\begin{bmatrix} x\\ y\\ z \end{bmatrix}=\begin{bmatrix} d_2^2-(x_2^2+y_2^2+z_2^2)-d_1^2+(x_1^2+y_1^2+z_1^2)\\ d_3^2-(x_3^2+y_3^2+z_3^2)-d_2^2+(x_2^2+y_2^2+z_2^2)\\ d_4^2-(x_4^2+y_4^2+z_4^2)-d_3^2+(x_3^2+y_3^2+z_3^2)\\ d_1^2-(x_1^2+y_1^2+z_1^2)-d_4^2+(x_4^2+y_4^2+z_4^2) \end{bmatrix} \quad (6)$$

Represent the equation in form $2AX = B$, now the X coordinate value is calculated as follows:

$$X = \frac{1}{2}(A^T A)^{-1} A^T B \quad (7)$$

TABLE II

| Test Point | Actual Position (m) | Mean Error (cm) | Max Error (cm) | RMSE (cm) |
|---|---|---|---|---|
| P1 | (0.5, 0.5) | 8.2 | 14.5 | 9.1 |
| P2 | (2.5, 1.5) | 11.4 | 19.3 | 12.7 |
| P3 | (3.75, 3.75) | 9.8 | 16.2 | 10.3 |
| P4 | (1.5, 5.0) | 12.6 | 21.4 | 13.9 |
| P5 | (5.0, 2.0) | 10.1 | 17.8 | 11.2 |
| Average | | 10.9 | 18.6 | 11.9 |

To quantitatively evaluate the localization performance of the UWB subsystem, measurements were conducted at five predefined reference points distributed across the experimental area. At each point, 50 consecutive readings were recorded and compared against the ground-truth coordinates. Table II summarizes the mean error, maximum error, and root mean square error (RMSE) obtained at each location, demonstrating that the system achieves an average RMSE of 11.9 cm, which is sufficient for reliable pedestrian-level navigation guidance.

## 2.2 Build grid maps from real environments

To facilitate path planning, the environment is discretized into a 2D occupancy grid map. In this model, each grid cell is assigned a binary value: 1 represents an obstacle, while 0 denotes navigable free space, as illustrated in Figure 3a. Figure 3b demonstrates the transformation of a physical environment into a mesh representation, which is essential for computing collision-free trajectories from the start to the goal. The grid cell resolution is carefully determined to balance spatial accuracy with the system's real-time computational efficiency.

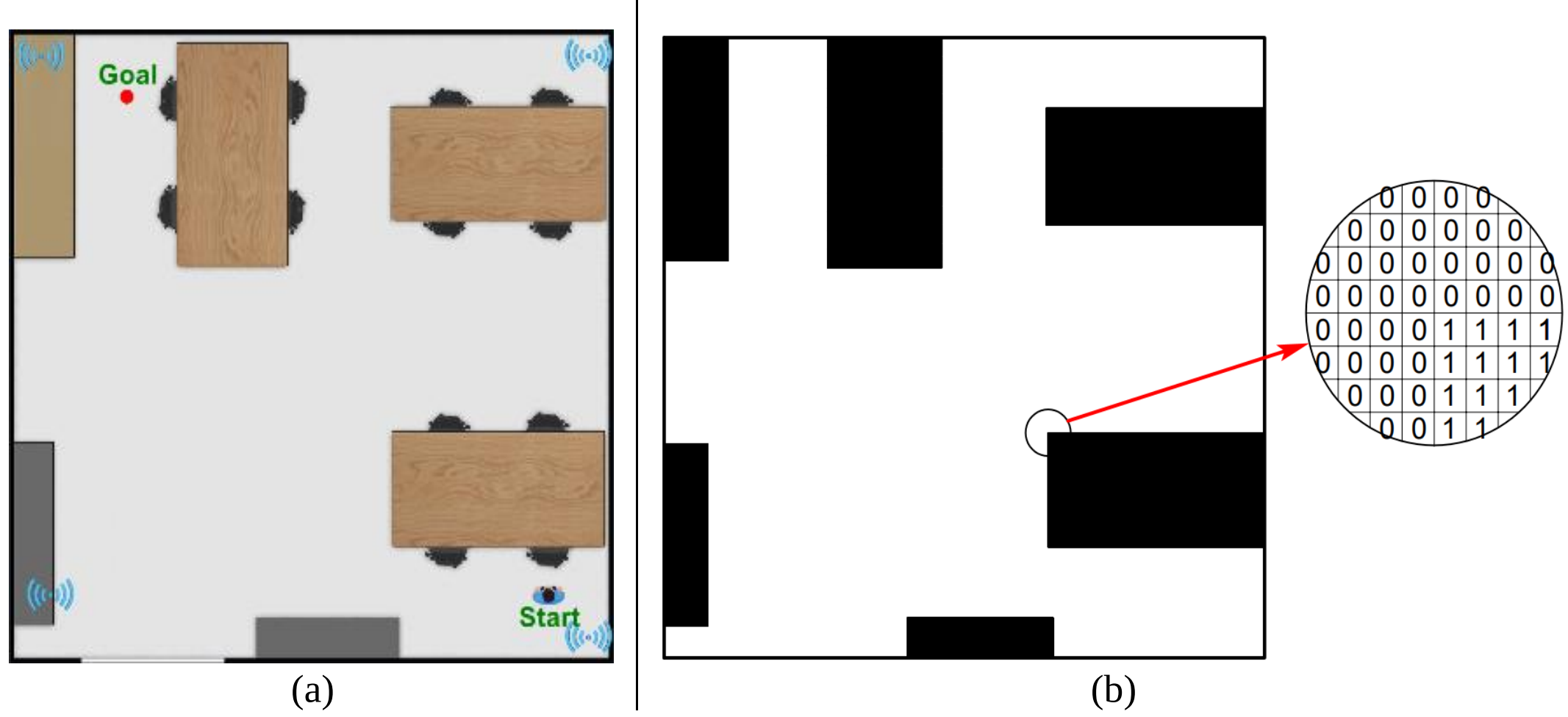

**Fig 3.** *Setting up the satellites in the experimental environment*

## 2.3 Finding part using the D* Lite algorithm

The D* Lite algorithm is an incremental heuristic search method derived from Lifelong Planning A* (LPA*) and the foundational A* algorithm [15][16][17][21]. A core advancement in D* Lite is the introduction of the right-hand side (*rhs*) variable, which serves as a lookahead estimate of the *g*-value. The algorithm excels in incremental path planning by treating unknown regions as free space and performing a backward search (from the target toward the start), distinguishing it from A* and LPA* [18][19][20][22].

In each iteration, D* Lite expands eight neighboring nodes by selecting the minimum *rhs* value until the target is reached. Here, *g(s)* and *rhs(s)* represent the cost estimates from node *s* to the goal, where *rhs(s)* is computed based on the *g*-values of its successors as follows:

$$rhs(s) = \begin{cases} 0, & s = s_{goal} \\ \min\limits_{s' \in \mathrm{Pred}(s)} \left(g(s') + c(s', s)\right), & other \end{cases} \quad (8)$$

where $c(s', s)$ represents the edge cost of nodes $s'$ to $s$, usually recorded as 1. $s' \in \mathrm{Pred}(s)$ represents the set of nodes immediately preceding vertex $s \in S$. When $g(s) = rhs(s)$, it means the nodes are homogeneous, otherwise they are heterogeneous. Where, $g(s) > rhs(s)$ is called locally beyond uniformity, and $g(s) < rhs(s)$ is called locally inhomogeneous. When evaluating the estimated value of grid points, D* Lite also introduces the value $key(s)$ for comparison, where $key(s)$ contains two values $key(s) = [key(s_1); key(s_2)]$, satisfying the following formulas:

$$\begin{gathered} key1(s) = \min[g(s), rhs(s)] + h(s) \\ key2(s) = \min[g(s), rhs(s)] \end{gathered} \quad (9)$$

where, $k_1$ and $k_2$ are priority queue parameters. The size of the value $key(s)$ is used as the expansion priority, which determines the order in which nodes in the queue are expanded. First compare the $k_1$ value size, choose the grid with the smallest $k_1$ value as the next grid, and when the $k_1$ value is equal, consider the $k_2$ value size. The heuristics function $h(s)$ is similar to the heuristics function in the LPA* algorithm, representing the estimated value between the current node and the starting point, causing the object to always move towards the target point. The value of $h(s)$ is very important, it determines the speed and accuracy of the search algorithm. The search heuristics function of node s consists of two parts: the cost $c(s', s)$ from node s to node s' and the value of the heuristics function from node $s'$ to the goal point sgoal. The calculation formula of $h(s)$ is:

$$h(s, s_{start}) = \begin{cases} 0, & s = s_{goal} \\ c(s', s) + h(s', s_{goal}), & other \end{cases} \quad (10)$$

### 2.4 Voice guidance and Navigation logic

Once the system successfully determines the user's real-time pose—comprising both precise coordinates and orientation—and generates an optimal trajectory via the D* Lite algorithm, the Raspberry Pi 4 central processor initiates the audio guidance sequence. The navigation logic continuously monitors the angular deviation between the user's current heading and the desired path vector. Based on this discrepancy, the system dynamically triggers specific voice commands, such as 'Go straight,' 'Turn left,' 'Turn right,' 'Rotate left,' 'Rotate right,' or 'Arrived at goal.' These commands are executed only when the deviation exceeds pre-defined angular thresholds, ensuring the guidance is both accurate and non-intrusive. This precise mapping of angular ranges to specific audio feedback is illustrated in Figure 5, while the comprehensive operational logic and the detailed voice-instruction flowchart are provided in Figures 4 and 6b, respectively, highlighting the system's robust decision-making framework.

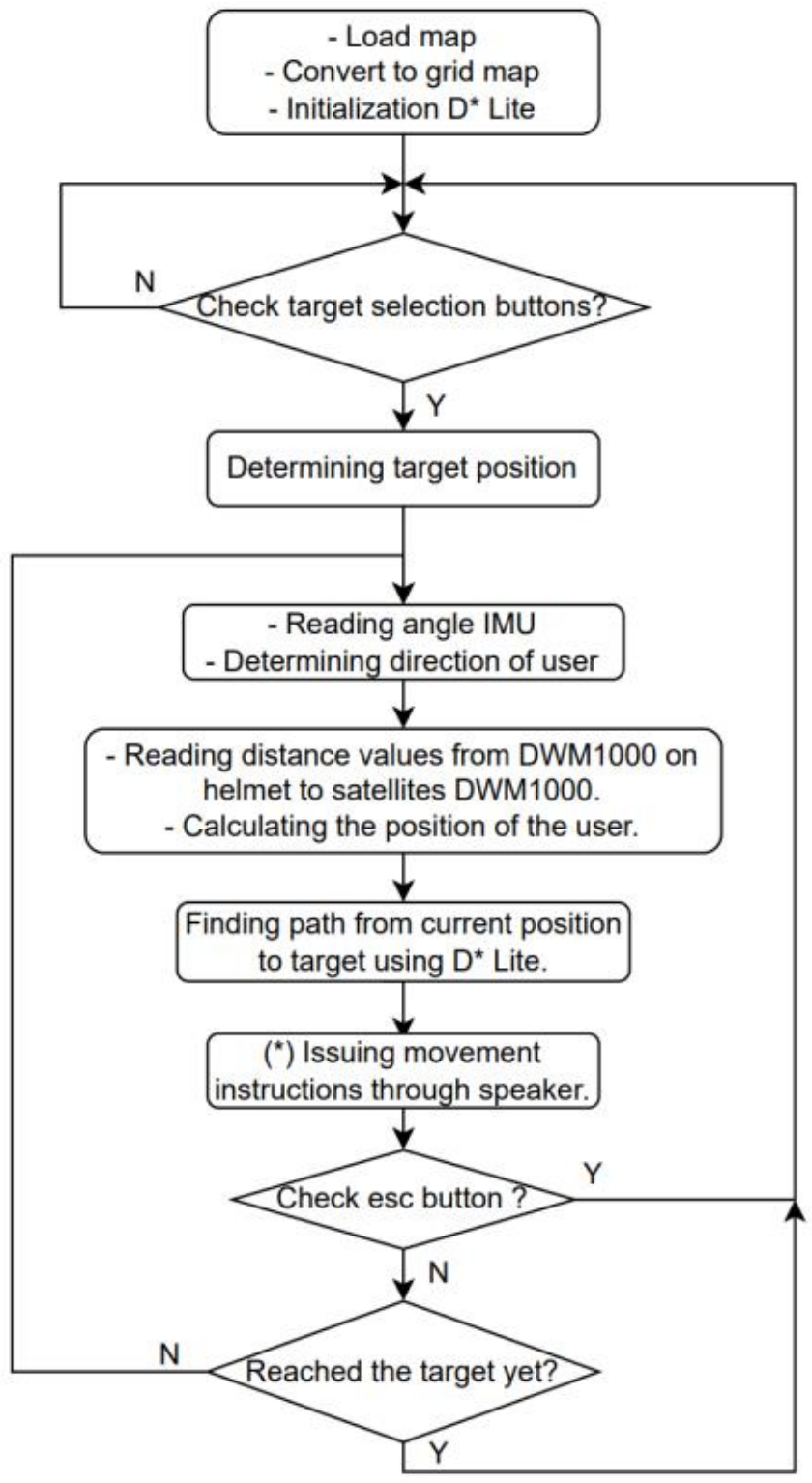

**Fig 4.** *Main programming algorithm*

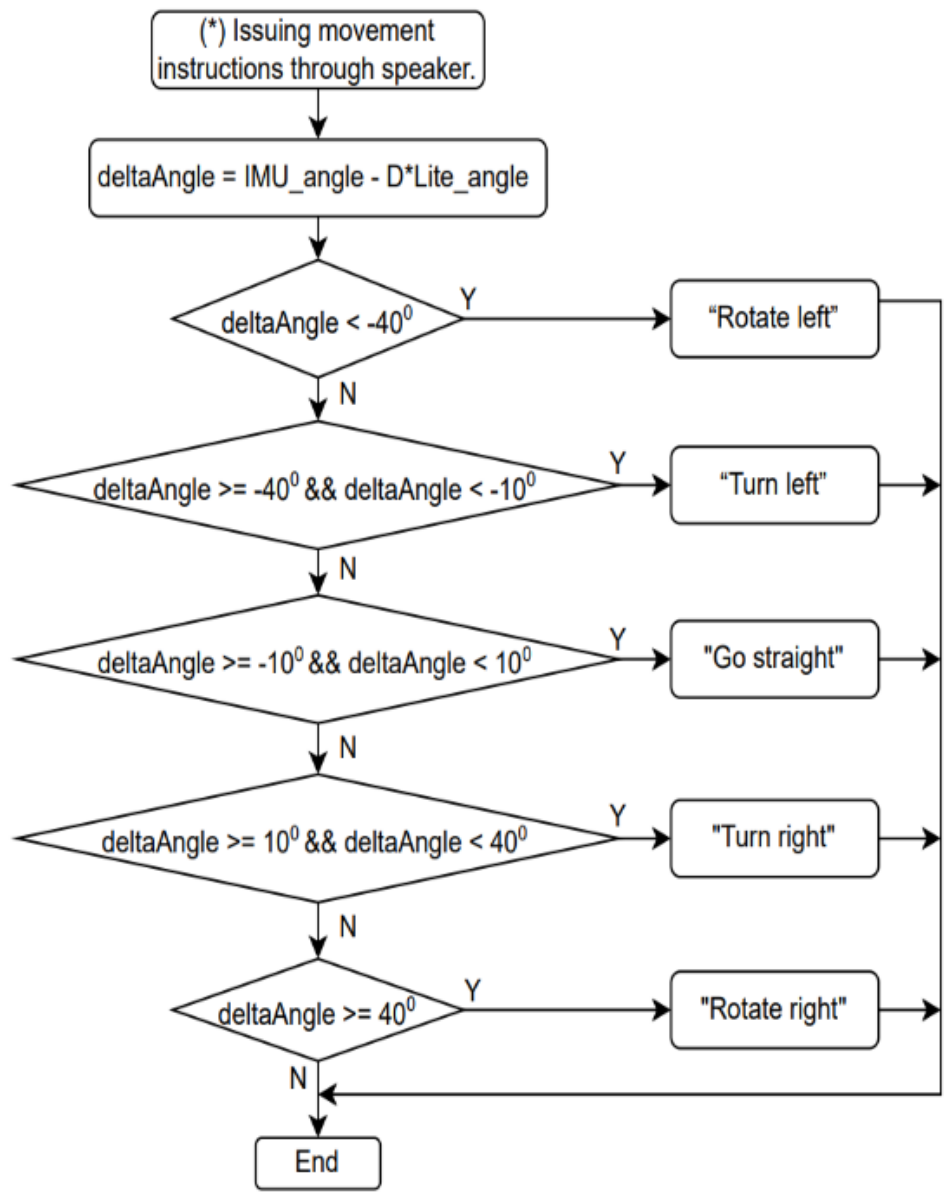

**Fig 5.** *Algorithm for issuing voice instructions*

## 3. Experimental results

The experimental setup was conducted in a 7.5x7.5m indoor environment containing fixed obstacles such as tables, chairs, and cabinets. Figure 6 shows a prototype consisting of a helmet with integrated sensors and a processor. Figure 7 illustrates the system interface, displaying the real-time positions of the user, the UWB anchors, the discretized obstacles, and the selected destination. A grid map with a resolution of 0.1m per cell was pre-constructed and integrated into the software, with anchor coordinates precisely mapped to their physical locations.

Upon system initialization, an audible 'beep' notifies the user that navigation is ready. The interaction is managed via head-mounted buttons: one for cycling through pre-defined destinations—confirmed by audio feedback—and another for initiating the navigation process upon a second press. Additionally, a dedicated button allows the user to terminate the current session and re-select a destination. The experimental results, showcased in Figure 7, depict the user's trajectory and real-time state, with corresponding voice commands displayed at the header of each frame.

To further assess the reliability of the complete navigation pipeline, end-to-end trials were carried out under two distinct environmental conditions: a static setup with fixed furniture and a reconfigured layout simulating real-world changes. Table III reports the number of trials, successful navigations, and the corresponding success rates for each scenario. The system achieved an overall success rate of 91.4%, with a slight reduction observed in the rearranged layout condition, which is attributed to accumulated localization drift near furniture clusters acting as signal reflectors.

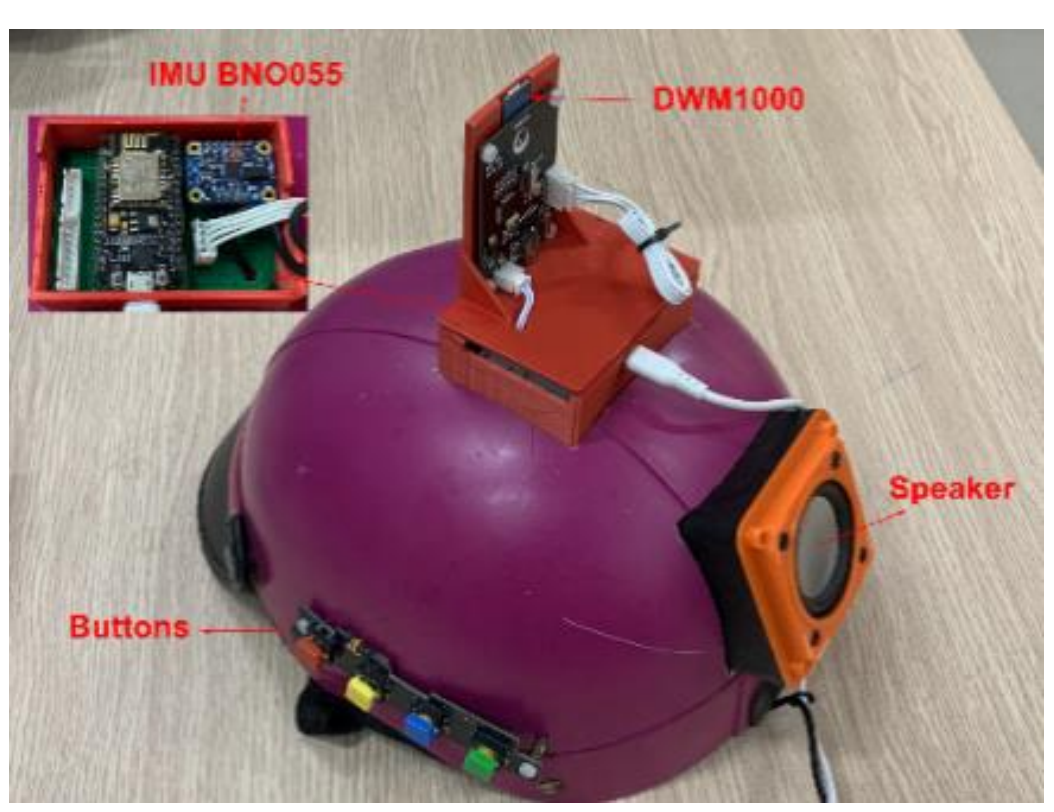

**Fig 6.** *Prototype consisting of a helmet with integrated sensors and a processor*

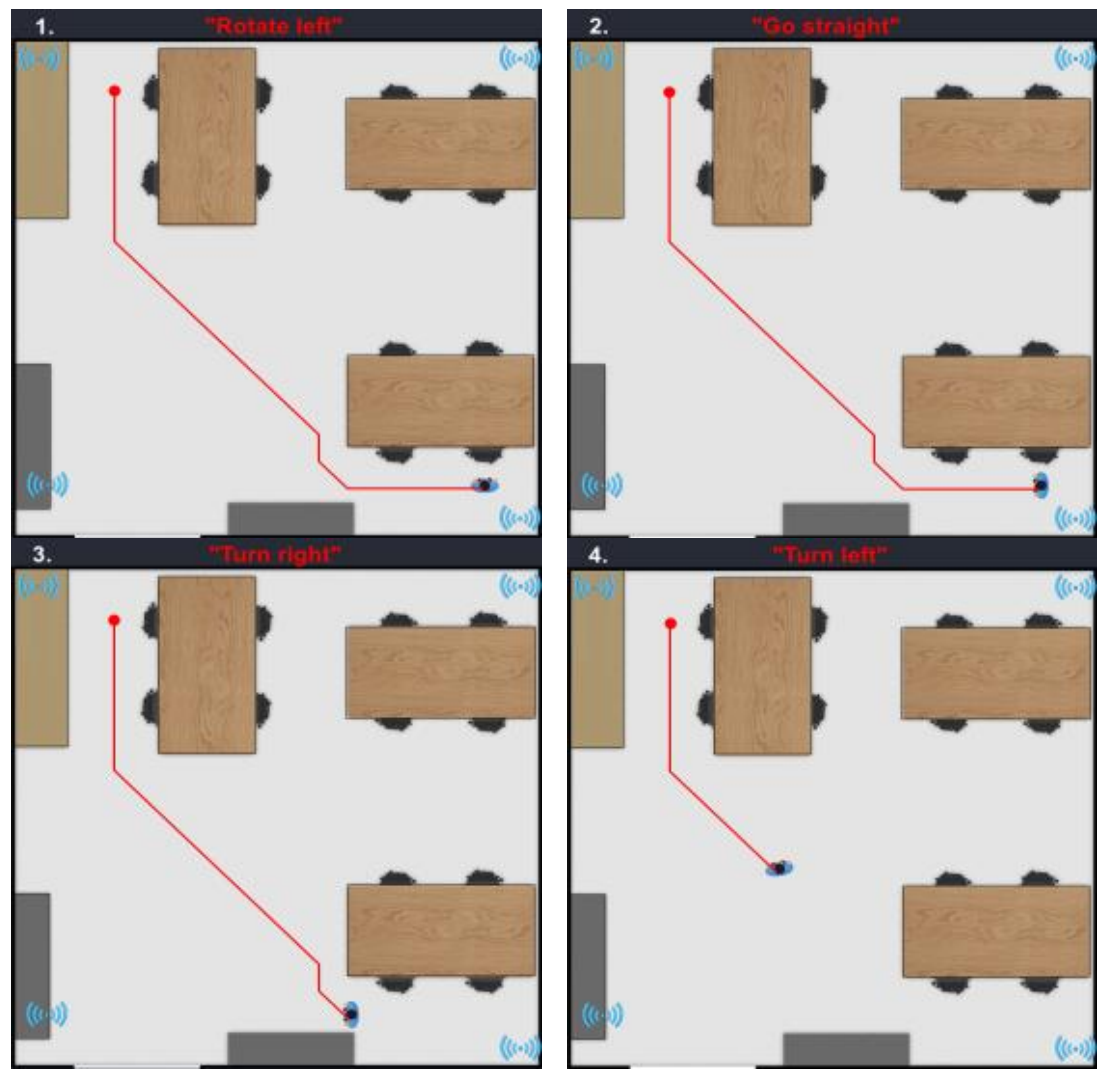

**Fig 7.** *The results of voice-guided navigation*

TABLE III

| Scenario | No. of Trials | Successful | Failed | Success Rate (%) |
|---|---|---|---|---|
| Static environment | 20 | 19 | 1 | 95.0 |
| Rearranged furniture layout | 15 | 13 | 2 | 86.7 |
| Total | 35 | 32 | 3 | 91.4 |

## 4. Conclusions and Future work

This study has successfully developed an innovative head-mounted assistive device to facilitate indoor localization and navigation for the visually impaired. Experimental results demonstrate that the integration of the BNO055 IMU and DWM1000 modules provides high-precision orientation and positioning, enabling users to navigate unfamiliar environments with ease. Furthermore, the implementation of the D^* Lite algorithm ensures optimal path planning and effective avoidance of static obstacles such as furniture and walls.

The proposed solution significantly enhances the independence and confidence of visually impaired individuals in daily activities, while contributing a robust framework to the field of indoor positioning technology. Future research will focus on expanding the system's operational scope, enhancing computational efficiency, and integrating advanced features for dynamic obstacle detection and avoidance. Such advancements aim to provide more practical and comprehensive mobility solutions for the visually impaired and broader disabled communities.